\documentclass[letterpaper]{article} % DO NOT CHANGE THIS
\usepackage{aaai2026}
\usepackage{times}
\usepackage{helvet}
\usepackage{courier}
\usepackage{xcolor}
\usepackage{aaai2026}  % DO NOT CHANGE THIS
\usepackage{times}  % DO NOT CHANGE THIS
\usepackage{helvet}  % DO NOT CHANGE THIS
\usepackage{courier}  % DO NOT CHANGE THIS
\usepackage[hyphens]{url}  % DO NOT CHANGE THIS
\usepackage{graphicx} % DO NOT CHANGE THIS
\usepackage{enumitem}
\usepackage{subcaption}
\usepackage{amsthm}
\usepackage{amsmath}
\usepackage[table]{xcolor}
\usepackage{amssymb}
\usepackage{booktabs}
\usepackage{multirow}
\usepackage{enumitem}

\usepackage{bm}
\usepackage{xcolor}
\usepackage{caption} 
\usepackage{amsfonts}
\usepackage{amsmath}
\usepackage{amssymb}
\usepackage{natbib}  % DO NOT CHANGE THIS AND DO NOT ADD ANY OPTIONS TO IT
\usepackage{caption} % DO NOT CHANGE THIS AND DO NOT ADD ANY OPTIONS TO IT
\usepackage{algorithm}
\usepackage{algorithmic}

\usepackage{newfloat}
\usepackage{listings}
\DeclareCaptionStyle{ruled}{labelfont=normalfont,labelsep=colon,strut=off} % DO NOT CHANGE THIS
\floatstyle{ruled}
\newfloat{listing}{tb}{lst}{}
\floatname{listing}{Listing}
\title{Out-of-Distribution Detection with Positive and Negative Prompt Supervision Using Large Language Models}

\author{
    Zhixia He$^1$, Chen Zhao$^2$, Minglai Shao$^{1}\textsuperscript{*}$, Xintao Wu$^{3}$, Xujiang Zhao$^{4}$, Dong Li$^2$, Qin Tian$^{5}$, Linlin Yu$^{6}$
    }

\affiliations{
    \textit{\textsuperscript{\rm 1}School of New Media and Communication, Tianjin University, Tianjin, China\\
    \textsuperscript{\rm 2}Department of Computer Science, Baylor University, Waco, Texas, USA\\
    \textsuperscript{\rm 3}Department of Electrical Engineering and Computer Science, University of Arkansas, Fayetteville\\
    \textsuperscript{\rm 4}NEC Laboratories America\\
    \textsuperscript{\rm 5}College of Intelligence and Computing, Tianjin University, Tianjin, China\\
    \textsuperscript{\rm 6}Department of Computer Science, Augusta University, Augusta, GA, USA\\}
    \{2023245033, shaoml, tianqin123\}@tju.edu.cn, \{chen\_zhao, dong\_li1\}@baylor.edu, xintaowu@uark.edu, xuzhao@nec-labs.com, linyu@augusta.edu
    
}

\usepackage{bibentry}
\begin{document}

\maketitle
\newcommand{\sysname}{PNPS}
\begin{abstract}
Out-of-distribution (OOD) detection is committed to delineating the classification boundaries between in-distribution (ID) and OOD images. 
Recent advances in vision-language models (VLMs) have demonstrated remarkable OOD detection performance by integrating both visual and textual modalities.
In this context, negative prompts are introduced to emphasize the \textit{dissimilarity} between image features and prompt content. However, these prompts often include a broad range of non-ID features, which may result in suboptimal outcomes due to the capture of overlapping or misleading information. To address this issue, we propose \textit{Positive and Negative Prompt Supervision}, which encourages negative prompts to capture inter-class features and transfers this semantic knowledge to the visual modality to enhance OOD detection performance. Our method begins with class-specific positive and negative prompts initialized by large language models (LLMs). These prompts are subsequently optimized, with positive prompts focusing on features within each class, while negative prompts highlight features around category boundaries. Additionally, a graph-based architecture is employed to aggregate semantic-aware supervision from the optimized prompt representations and propagate it to the visual branch, thereby enhancing the performance of the energy-based OOD detector. Extensive experiments on two benchmarks, \texttt{CIFAR-100} and \texttt{ImageNet-1K}, across eight OOD datasets and five different LLMs, demonstrate that our method outperforms state-of-the-art baselines.
\end{abstract}
\section{Introduction}
When deploying machine learning models in open-world scenarios, it is inevitable to encounter samples from previously unseen classes, commonly referred to as \textit{out-of-distribution} (OOD) data \cite{hendrycks2016baseline,shao2024supervised,lin2024fade,zhao2021fairness,wu2025explainable,lin2025face4fairshifts}. This issue is particularly critical in high stakes applications, such as autonomous driving \cite{bogdoll2022anomaly} and medical diagnostics \cite{li2025multi,zimmerer2022mood,li2025survey}, where the misclassification of OOD samples can pose significant safety hazards. Moreover, recent studies have revealed that even state-of-the-art deep neural networks often make overconfident predictions on OOD data \cite{parmar2023open,lid2024learning}. Consequently, there is a growing need for OOD detection methods that can effectively identify OOD samples.

\begin{figure}[t!]
\centering
\captionsetup{skip=5pt}
\includegraphics[width=1\linewidth]{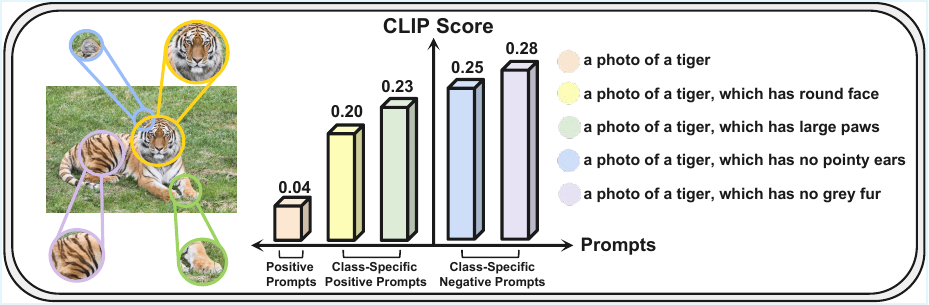}
\caption{\textbf{Illustration of the CLIP score using prompts and an image of a tiger.} Compared to the prompt  ``\textit{a photo of a tiger}'', positive prompts enriched with visual features significantly enhances the CLIP score. Furthermore, the introduction of negative features can further increase  the score. }
\label{picture_1}
\vspace{-0.65cm}
\end{figure}

Traditional OOD detection methods primarily rely on a single visual modality, overlooking the rich semantic content carried by labels \cite{liu2020energy}.  With the emergence of vision-language models (VLMs),  a paradigm shift has occurred---from relying on vision-only information to integrating both visual and textual modalities \cite{zhang2024visionlanguagemodelsvisiontasks}. In the context of VLM-based OOD detection, the textual modality is typically represented by positive prompts (e.g., \textit{``a photo of a \{class\}"}), which are used to estimate the probability of an image belonging to the given class.
However, these fixed-format prompts do not account for the distinctions between categories.  
To address this challenge, class-specific positive prompts enriched with visual features generated by large language models (LLMs) have been introduced, demonstrating superior performance compared to those using only category names \cite{menon2022visual}.
More recently, CLIPN \cite{wang2023clipn} incorporates \textit{``no class''} into prompts to express negative concepts, which are referred to as negative prompts.
Despite these advances, the use of negative prompts may still result in the learning of overlapping non-ID features or noisy information, potentially leading to suboptimal performance. Given the limitations of existing endeavors, our twofold objective is to: (1) encourage positive and negative prompts to comprehensively capture ID category features and to clearly delineate the category boundaries; and (2)  aggregate semantic-aware supervision from prompt representations and transfer it to the visual branch to improve OOD detection.

One promising direction is to construct class-specific negative prompts by augmenting positive prompts with negative features. To preliminarily explore its effectiveness, we select an image of a tiger, prompt LLMs to generate class-specific descriptions, and calculate the corresponding CLIP score, as illustrated in Figure \ref{picture_1}.
The results indicate that, compared to positive prompts, class-specific negative prompts achieve a better match with the image. As will be discussed later, these augmented prompts primarily capture inter-class features and guide the model focus on category distinctions, thus better matching the corresponding image. In this paper, we propose the  \textbf{P}ositive
and \textbf{N}egative \textbf{P}rompt \textbf{S}upervision (\textbf{{\sysname}}) framework. This framework consists of three phases:  prompt construction with large language models, alignment of visual and textual modalities, and cross-modal graph neural networks. We first construct positive and negative prompts by prompting LLMs to generate class-specific visual descriptions. To further enhance the expressiveness of these prompts, we introduce learnable textual parameter matrices, which enable positive and negative prompts to effectively capture intra-class and inter-class features, respectively. However, a key challenge for improving OOD detection lies in how to aggregate semantic-aware supervision from the optimized prompts and propagate it to the image representations. To address this, we build multi-modal graphs to facilitate the aggregation and propagation of semantic supervision within and across modalities. Extensive experiments on two ID and eight OOD datasets using five different LLMs demonstrate the effectiveness of our proposed {\sysname} method.
Our main contributions are:
\begin{itemize}[leftmargin=*]
\item We employ a graph-based structure to aggregate both positive and negative prompt representations as semantic-aware supervision, which is then propagated to the visual modality to enhance OOD detection performance.  To our knowledge, we are the first to utilize a graph-based framework for OOD detection in the image domain.

\item 
We introduce a novel three-phase \sysname{} framework that optimizes class-specific positive and negative prompts to capture intra-class and inter-class features, facilitating a more profound understanding of ID features as well as the delineation of more clearly defined category boundaries.

\item 
On the \texttt{CIFAR-100} benchmark, our {\sysname} improves AUROC by $1.06\%$, $2.10\%$, $4.41\%$, and $3.97\%$ over the best baseline on the \texttt{CIFAR-10}, \texttt{SVHN}, \texttt{Texture}, and \texttt{Places365}, respectively. In addition, our approach also achieves superior performance on \texttt{ImageNet-1K}.

\end{itemize}

\section{Related Work}
\textbf{Out-of-Distribution Detection.} OOD detection aims to identify images that do not belong to any category in the training dataset. Early research primarily focuses on designing score functions based on the predicted logits, such as the energy score \cite{liu2020energy}. Data augmentation techniques are also adopted to enhance data diversity and improve model generalization, by applying effective transformations to the training data \cite{goodfellow2020generative, nie2024out}. Furthermore, later studies explore extracting class-agnostic information from the feature space, which is not accessible from predicted logits alone \cite{sun2021react}. For instance, ViM \cite{wang2022vim} enhances OOD detection performance by combining class-agnostic features with logits. Moreover, with recent advancements in VLMs, leveraging textual information becomes a new and promising direction to further improve image OOD detection.

\noindent\textbf{VLM-Based Out-of-Distribution Detection.} With the advancements in VLMs, a series of VLM-based OOD detection methods have rapidly emerged. As an early application, MCM \cite{ming2022delving} utilizes the maximum logit of scaled softmax to detect OOD images. Subsequently, the interaction between NLP and CV has facilitated the application of prompt learning from NLP to OOD detection, notably exemplified by CoOp \cite{zhou2022coop}.  
More recently, with the emergence of LLMs, prompt-based LLMs have enabled the automatic generation of class-specific visual features. Methods such as DCLIP \cite{menon2022visual} and CuPL \cite{pratt2023does} demonstrate that prompts with detailed descriptions can further enhance the matching between images and prompts.  Building on these studies, a promising direction involves simulating non-ID scenarios to improve OOD detection performance. 
For instance, CLIPN \cite{wang2023clipn} introduces negative prompts, such as \textit{`` a photo of no \{class\}"}, to capture negation semantics within images, while EOE \cite{cao2024envisioning} leverages LLMs to generate pseudo-OOD labels. Although the aforementioned methods have shown promising results, explorations that combine prompts with negative features remain limited, leaving significant potential for further research.

\section{Preliminaries}
Let $\mathcal X$ denotes the image space and $\mathcal Y = \{y_1,y_2,...,y_{|\mathcal C|}\}$ represents the set of ID class labels, where $|\mathcal C|$ is the total number of classes. We define $x$ as the random variable sampled from $\mathcal X$, with each sample associated with a label \(y \in \mathcal Y\). Notably, the training and testing data are drawn from different distributions, \textit{i.e.}, \(\mathbb P^{tr}(\mathcal{X}, \mathcal{Y})\ne \mathbb P^{te}(\mathcal{X}, \mathcal{Y})\), where test set may include OOD instances with labels $y \notin \mathcal Y$. 

\noindent \textbf{CLIP and DCLIP.} CLIP \cite{radford2021learning} is trained on 400 million image-text pairs and utilizes contrastive learning to align visual and textual representations.
It comprises a text encoder $\mathcal T:t \rightarrow \mathcal R^d $ and an image encoder $\mathcal I:x \rightarrow \mathcal R^d$. During inference, given an image-label pair  $(x,y)\in(\mathcal X,\mathcal Y)$, a prompt like \textit{``a  photo  of  a \{$y$\}"} is fed to the text encoder to obtain $\mathcal T(t)$. The predicted probability for the image feature $\mathcal I(x)$ is then computed as follows:
\begin{equation}
    p(y=i|x) = \frac{e^{<\mathcal I(x), \mathcal T(t^i)>/\tau}}{\sum_{c=1}^{|{\mathcal C}|} e^{<\mathcal I(x), \mathcal T(t^c)>/\tau}},
\end{equation}
where $\tau$ is the temperature parameter. The following introduces DCLIP \cite{menon2022visual}, which leverages LLMs to generate visual features that describe the object category in a photo. Specifically, it prompts the LLMs with the following query \textit{\textbf{Q}} and obtains the corresponding answer \textit{\textbf{A}}:
\begin{center}
\textit{\textbf{Q}: What are useful features for distinguishing a \{class\}}\\
\textit{ in a photo?}\\
\textit{\textbf{A}: There are \{visual features\} to tell there is a \{class\} }\\
\textit{in a photo.}
\end{center}

The generated visual features are embedded into a fixed-format template to construct class-specific positive prompts, which outperform those using only the category names.  For fair comparison with the template \textit{``a photo of a \{class\}"}, we standardize the class-specific positive prompt format as \textit{``a photo of a \{class\}, which has \{visual features\}"}.

\section{Methodology}
In this section, we describe our proposed  {\sysname} framework. As shown in Figure \ref{large_picture}, the architecture consists of three main phases: 1) utilizing LLMs to generate category-discriminative features for constructing prompts;  2) optimizing prompts to capture both inter-class and intra-class features; and 3) employing a graph-based structure to aggregate and propagate semantic knowledge extracted from prompts.
\subsection{Prompt Construction with Large Language Models }

Traditional OOD detection methods typically represent semantic categories  (e.g., tiger) as numerical labels (e.g., 0), which capture only index-based relationships and overlook rich semantic content. 
To bridge this gap, positive textual prompt templates,  such as  \textit{``a photo of a \{class\}"}, have been introduced; however, these prompts still fail to capture the intrinsic differences between categories. To emphasize the distinctions between categories, recent LLM-enriched approaches construct class-specific positive prompts by appending discriminative visual descriptions \cite{menon2022visual}. 
Likewise, negative prompts are introduced to learn the dissimilarity for each categories
 \cite{wang2023clipn}.

This naturally raises a question: could incorporating negative category features into prompt templates yield better performance? We refer to such prompts as \textit{``class-specific negative prompts"}, and evaluate their effectiveness with pre-trained CLIP based on ViT/B-16, as shown in Figure \ref{picture_1}. The results indicate that the use of these negative prompts leads to a better match with the image. Building on this finding, we further enhance the match by encouraging LLMs to generate more discriminative descriptions. 
Specifically, we  divide categories in the training set into super-classes based on species similarity, as exemplified by the five classes \textit{\{``tiger", ``wolf", ``bear", ``leopard", ``lion"\}} in the \texttt{CIFAR-100} dataset are grouped into  \textit{``large carnivores"} super-class. Take the tiger class as an example, we modify the original query from \textit{\textbf{Q}} to \textit{\textbf{Q'}} as follows: 
%for the LLMs as follows: 
\begin{center}
\textit{\textbf{Q'}: What are useful features for distinguishing a \{tiger\}}\\
\textit{from \{wolf, bear, leopard, lion\} in a photo?}
\end{center}

\begin{figure*}[t!]
\centering
\includegraphics[width=\linewidth]{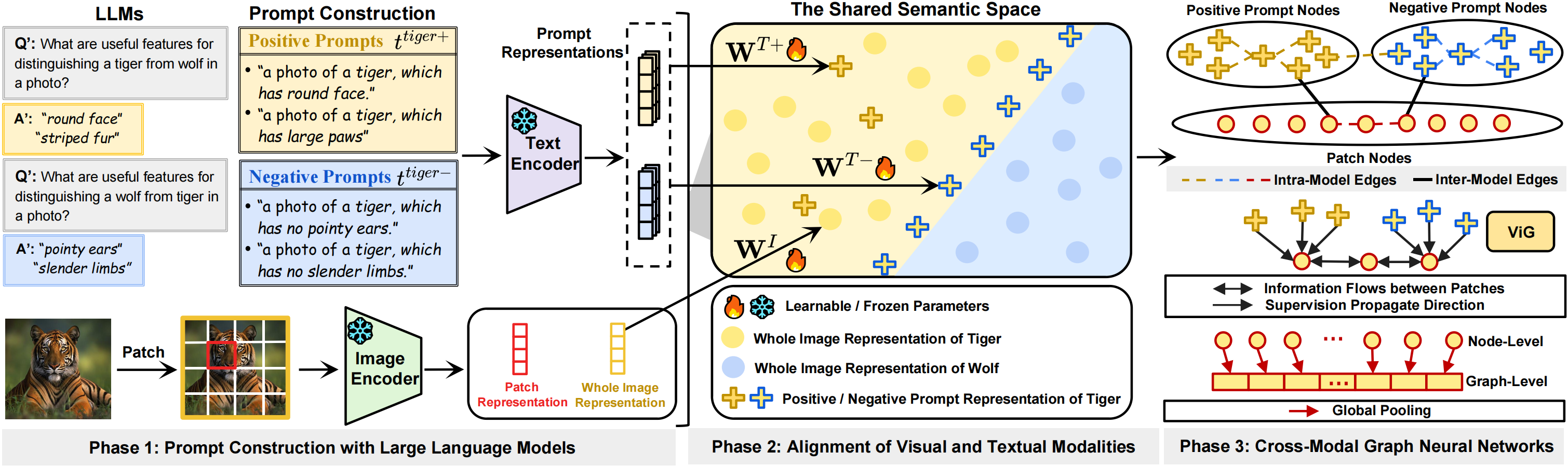}
\vspace{-5mm}
\caption{ An overview of the {\sysname} framework. To begin with, we employ LLMs to generate discriminative features, which are then filled into templates to construct class-specific positive and negative prompts. These prompts, together with image patches and full images, are subsequently encoded into their respective representations. To enhance the expressiveness of the prompt representations, we introduce learnable textual and visual parameter matrices,  $\textbf{W}^T$ and $\textbf{W}^I$, for further optimization. Building on these optimized textual representations, we then construct cross-modal graph connections to aggregate semantic supervision from the prompts and propagate it to the visual branch, thereby improving the performance of image OOD detection.}
\vspace{-4mm}
\label{large_picture}
\end{figure*}

The reason for the above is to encourage LLMs to generate non-overlapping features, thus offering more discriminative clues for category differentiation. As shown in Figure \ref{large_picture}, we incorporate the generated feature  \textit{``striped fur"}  into the template to construct the class-specific positive prompt  \textit{``a photo of a tiger, which has striped fur"}, symbolized  as ${t}^{tiger+}$. 
To exemplify, consider the task of distinguishing a tiger from a wolf, bear, leopard, and lion. Since \textit{``striped fur''} is unique to the tiger, its negation---\textit{``no striped fur"} serves as a complementary descriptor for the other four categories. By doing so, the model is guided to attend to fur-related features when distinguishing between categories. For instance, the class-specific negative prompt for the wolf is \textit{``a photo of a wolf, which has no striped fur"}, represented as ${ t}^{wolf-}$. 

All categories in the training dataset are grouped into $|\mathcal C^{super}|$ super-classes, where $\mathcal C^{super}$=\textit{\{``large carnivores",  ``trees", $\cdots$\}}. The super-class containing the tiger category is denoted as $\mathcal C^{super}_{tiger}$=\textit{\{``large carnivores"\}}. If $N$ visual features are generated for each category, then each category will have $N$ class-specific positive prompts and $(|\mathcal {C}^{super}_y|-1)\cdot N$ class-specific negative prompts. To illustrate, for the tiger category, negative prompts can be constructed by taking the negations of the $N$ visual features from each of the other four categories. These prompts are then fed into the text encoder $\mathcal T(\cdot)$ to obtain  representations $\textbf{H}^T=\{\textbf{H}^{T+}, \textbf{H}^{T-}\}$. 
\begin{equation}
\textbf{H}^T =\big\{\bm{h}_1^{T},\cdots,\bm{h}_N^{T},\bm{h}_{N+1}^{T},\cdots,\bm{h}_{|\mathcal{C}^{super}_y|\cdot N}^{T}\big\},
\end{equation}
where $\textbf{H}^{T+}\!\!=\!
\big\{\bm{h}_n^{T}\big\}_{n=1}^{N}$ and $\textbf{H}^{T-}\!\!=\!\big\{\bm{h}_{n}^{T}\big\}_{n=N+1}^{|\mathcal{C}^{super}_y|\cdot N}$. Ideally, positive prompts capture intra-class features, whereas negative prompts, constructed by combining a category name with the description of another class, should have their representations positioned at the boundaries between the two  categories.
However, due to the hallucination in LLMs-generated features, and the limited expressiveness of the pre-trained text encoder, the resulting representations may fail to convey the semantic meaning of \textit{``no”} \cite{nie2024out}. The subsequent step is to explore an effective method for optimizing the aforementioned prompt representations.

% While we understand that the negative meaning corresponds to "non-ID", we are still unclear about what specific features are being learned. 

\subsection{Alignment of Visual and Textual Modalities}
Inspired by  CLIP-Adapter  \cite{gao2021clip}, which achieves content optimization by adding an additional linear layer after the pre-trained text encoder in few-shot learning setting, we introduce a learnable prompt parameter matrix ${\textbf W}^T$, while keeping the pre-trained text encoder frozen. The resulting transformed textual representations are given by:
\begin{equation}
\hat{\textbf{H}}^{T}\!\!\!=\!\big\{\hat{\bm h}^{T}_1,\cdots\!,\hat{\bm h}^{T}_N,\hat{\bm h}^{T}_{N+1},\cdots\!,\hat{\bm h}^{T}_{|\mathcal C_y^{super}|\cdot N}\big\}\!=\!{\textbf W}^T\textbf{H}^{T}.
\end{equation}

To achieve better alignment between the visual and textual modalities, we also introduce a learnable visual parameter matrix ${\textbf W}^I$. 
The subsequent challenge lies in designing loss functions that enable class-specific positive prompts to effectively capture intra-class features, while negative prompts encode inter-class distinctions.  The optimization process involves three components: positive loss $\mathcal L^+$, negative loss $\mathcal L^-$, and the negative-positive distant loss $\mathcal L_{npd}$, which optimizes the relative distance between positive prompts and their corresponding negative prompts. Specifically, we begin by optimizing the positive representations as follows:

\subsubsection{Positive-Image Related Loss (PIR).}Aligning positive prompts with their respective images allows prompts to capture rich intra-class features. Specifically, for a given image $x$, we extract its representation $\bm{h}^I(x)$ with the pre-trained image encoder $\mathcal I(\cdot)$, and project it through the learnable  matrix $\textbf{W}^I$ to obtain the transformed image representation $\hat{\bm{h}}^I(x)$.  The prediction probability for an ID image $x$ and the $i$-th positive prompts $ t^{i+}$  is then computed as follows:
\begin{equation}
p^+_i = \frac{\sum_{n=1}^{N}e^{<\hat{\bm{h}}^I(x),\hat{\bm{h}}^{T}_n(t^{i+})>/\tau}}{\sum_{c=1}^{|\mathcal {C}|}\sum_{n=1}^{N} e^{<\hat{\bm{h}}^I(x),\hat{\bm{h}}^{T}_n(t^{c+})>/\tau}},
\end{equation}
\noindent where $|\mathcal {C}|$ denotes the total number of ID classes, and $N$ represents the number of positive prompts  per category. The positive-image related loss $\mathcal L_{pir}$ is defined as:
\begin{equation}
\mathcal L_{pir} = -\mathbb E_{{(x,t^{i+})}\thicksim \mathcal D^{tr}}
\log p^+_i.
\end{equation}

\subsubsection{Positive-Positive Distant Loss (PPD).}The positive-positive distant loss $\mathcal L_{ppd}$ is introduced to encourage diversity among the $N$ class-specific positive representations and ensure that each one captures distinct semantic information.
\begin{equation}
\mathcal L_{ppd} =  \sum_{c=1}^{|\mathcal C|} \sum_{i=1}^N \sum_{j=i+1}^N |<\hat{\bm{h}}^{T}_i(t^{c+}),\hat{\bm{h}}^{T}_j(t^{c+})>|.
\end{equation}

The overall objective for optimizing positive representations is defined as  $\mathcal{L}^+ = \mathcal{L}_{pir} + \lambda^+ \cdot\mathcal{L}_{ppd}$, with  $\lambda^+$ serving as a hyperparameter to balance the two components. The optimization process for negative prompts is described below:

\subsubsection{Negative-Image Related  Loss (NIR).}
We denote class-specific negative template \textit{``a photo of  a }\{{$y_{a}$}\}, \textit{which  has \{no visual features of  $y_{{b}}$\}"} as $t^{{a}-{b}}$, where $y_{a}, y_{b} \in \mathcal C_{[y_{ a},y_{ b}]}^{super}$ and the negative features of  $y_{b}$ serve as the complementary descriptor for $y_{a}$.  Accordingly, the representation $\hat{{\bm h}}^{T}(t^{{a}-{b}})$ should be aligned with the image representations of  $y_{a}$,  and diverge from those of  $y_{b}$ in the shared semantic space. This design facilitates the learning of well-defined category boundaries between $y_a$ and $y_b$, and enables the model to capture more discriminative inter-class features for $y_a$.
The matching probability between  $x$ and the $i$-th negative prompts $t^{i-}$ is computed as follows:
\begin{equation}
p^-_i=\sum_{\substack{c=1, c \neq i}}^{|\mathcal{C}_{[i,c]}^{super}|}\sum_{n=c \cdot N+1}^{(c+1)\cdot N+1}s^-_i(c,n),
\end{equation}
where $s^-_i(c,n)$ is the matching score between image $x$ and the $n$-th negative prompt of the $c$-th category within the coexisting super-class, formulated as follows:
\begin{equation}
s^-_i(c,n)=
\frac{e^{<\hat{\bm{h}}^I(x),\hat{\bm{h}}^{T}_{n}(t^{i-c})>}}{ e^{<\hat{\bm{h}}^I(x),\hat{\bm{h}}^{T}_{n}(t^{i-c})>} + e^{<\hat{\bm{h}}^I(x),\hat{\bm{h}}^{T}_{n}(t^{c-i})>}},
\end{equation} 
where the negative representations for class $y_c$ are indexed from {\small $c\cdot N+1$} to {\small$(c+1)\cdot N+1$}. $\mathcal L_{nir}$ is computed as follows:
\begin{equation}
\mathcal L_{nir} = -\mathbb E_{{(x,t^{i-})}\thicksim \mathcal D^{tr}} {\small\frac{1}{(|\mathcal{C}_i^{super}|-1) \cdot N }}
\log p^-_i.
\end{equation}

\subsubsection{Negative-Negative Distant Loss (NND).} To ensure diversity and non-overlap among negative prompts, we enforce greater separation between distinct negative representations.
\begin{equation}
\mathcal L_{nnd}\!=\!\!\!\sum_{c=1}^{|\mathcal C|}\!\!\sum_{\substack{d=1 \\ d \neq c}}^{|\mathcal C_{[c,d]}^{super}\!|}\!\sum_{\substack{i=c \cdot \\ N+1}}^{\substack{(c+1) \\ \cdot N}}\!\sum_{\substack{j= \\ i+1}}^{\substack{(c+1) \\ \cdot N}}
\!\!|\!\!<\!\hat{\bm {h}}_{i}^{T}(t^{c-d}),\hat{\bm {h}}_{j}^{T}(t^{c-d})\!\!>\!\!|.
\end{equation}

The total loss for negative representations is formulated as $\mathcal L^{-} = \mathcal L_{nir} + \lambda^{-} \cdot \mathcal L_{nnd}$. In addition to the  positive loss $\mathcal L^{+}$ and the negative loss $\mathcal L^{-}$, we introduce negative-positive distance loss $\mathcal L_{npd}$ to increase the separation between positive and their respective negative prompt representations. 

%After training, the learned negative representations place greater emphasis on inter-class distinctions while reinforcing category distinctiveness and diversity.
\subsubsection{Negative-Positive Distant Loss (NPD).} Theoretically,  the positive prompt ${t}^{a+}$ and the negative prompt ${t}^{b-a}$ are semantically opposite in the shared semantic space. To further enforce their distinction, we define $\mathcal L_{npd}$ as follows:
\begin{equation}
\mathcal L_{npd} \!= \!\! \sum_{c=1}^{|\mathcal C|}\!\sum_{\substack{d=1 \\ d \neq c}}^{|\mathcal C_{[c,d]}^{super}|}\!\sum_{n=1}^{N} |\!<\!\hat{\bm {h}}^{T}_n(t^{c+}), \hat{\bm {h}}_{d \cdot N+n}^{T}(t^{d-c})\!>\!\!|,
\end{equation}
where for the positive representation $\hat{\bm{h}}_{n}^T(t^{c+})$, the corresponding negative index is $d \cdot N+n$ for $d\in \mathcal C_{[c,d]}^{super}$. Once optimized, the prompts are enriched with semantic-aware information. The next objective is to transfer this semantic supervision to the visual branch to enhance OOD detection.

\subsection{Cross-Modal Graph Neural Networks}
\subsubsection{Multi-Modal Graph Construction.} Graph-based architecture offers a natural solution for aggregating semantic supervision from the optimized prompt representations and propagating it to the visual branch. Heterogeneous graphs, in particular, facilitate bidirectional message passing and feature updates between different node types \cite{zhang2019heterogeneous}. Within this structure, information flows via intra-modal and inter-modal edges within the multi-modal graph.

As class-specific prompts incorporate local detail descriptions,
the model is guided to attend to semantically consistent image features during optimization.
Consequently, the optimized prompt representations are more concentrated on these local features. To align with this locality, we employ the pre-trained ViT to extract patch features and link them to prompts via cross-modal edges.
Specifically, for an image $x$ and its corresponding label $y$, we obtain $|\mathcal C_y^{super}|\cdot N$ prompt representations and $M$  patch  representations. To facilitate this connection, we first construct an unordered node set $\mathcal{V}=\{\mathcal V^P, \mathcal V^T\}$, comprising patch and prompt nodes.
\begin{equation}
\mathcal V =\big\{v_1^P, \cdots, v_M^P, v_{M+1}^T, \cdots,
v_{|\mathcal C_y^{super}| \cdot N + M}^T\big\}.
\end{equation}
%We then establish intra-modal edge $e_{ij}^m$ by connecting $v_i^m$ with its neighbor $v_j^m \in \mathcal N(v_i^m)$.

For each node $v_i \in \mathcal V^m$ of the modality {$m\in \{P,T\}$}, we identify its $K^m$ nearest intra-modal neighbors as $\mathcal N(v_i^m)$. 
\begin{equation}
\mathcal N\!(v_i^m)\!=\!\!\big\{v_j\!\in \!\mathcal V^m|{i\neq j},v_j \!\in  \!\text{Top}K^m\!\big[\text{sim}(v_i,v_j)\big]\!\big\}.
\end{equation}

To lay the foundation for the construction of inter-modal edges, we also identify inter-modal neighbor set $\mathcal N(v_i^M)$.
\begin{equation}
\begin{aligned}
\mathcal{N}\!(v_i^M)\!\!= &\left\{\! v_j \!\in\! \mathcal{V}^T|v_i \!\in \!\mathcal{V}^P\!\!,\! v_j \!\in\! \text{Top}K^{\!M}\!\big[\text{sim}(v_i, \!v_j)\!\big] \!\right\} \\
\cup &\left\{ \!v_j \!\in\! \mathcal{V}^P| v_i \!\in\! \mathcal{V}^T\!\!,\!v_j \!\in\! \text{Top}K^{\!M}\!\big[\text{sim}(v_i, \!v_j)\!\big] \!\right\}\!,
\end{aligned}
\end{equation}
where we use the Euclidean distance as a measure of similarity. Based on the constructed neighbor sets, we then construct the intra-modal edge set $\mathcal{E}^{intra}=\{\mathcal{E}^P, \mathcal{E}^T\}$, and the inter-modal edge set  $\mathcal{E}^{inter}$. Building upon these structures, we construct the multi-modal graph $ G=(\mathcal V, \mathcal E)$, where $\mathcal V$, $\mathcal E=\{ \mathcal{E}^{intra}, \mathcal{E}^{inter}\}$ represent the set of nodes and edges.  

\subsubsection{Multi-Modal Graph Representation Learning.} Vision GNN (ViG) \cite{han2022vision} is proposed to effectively capture the relationships between patches.  In particular, graph-based architectures excel at modeling complex interactions among different types of nodes. Therefore, employing ViG to model intra-modal and inter-modal relationships holds significant potential. Given $M$ patch representations and and $|\mathcal C_y^{super}|\cdot N$ optimized prompt representations,  we employ  ViG to obtain the aggregated node-level representations.
\begin{equation}
\textbf H^{node} \!\!=\! \!\text{ViG}(G)\!=\!\{\bm{h}_1,\cdots\!,\bm{h}_M,\cdots\!,\bm{h}_{|\mathcal C_y^{super}|\cdot N+M}\!\}.
\end{equation}

Semantic supervision, aggregated from optimized prompt
representations, is propagated to the visual branch.
To analyze the impact of visual representations on OOD detection, we slice the node-level output to get patch-related components $\{\bm{h}_i\}_{i=1}^M$. A global pooling operation is then applied to convert these components into graph-level representation.
\begin{equation}
\textbf H^{global}=\text{Pooling}({\bm h}_1,{\bm h}_2,\cdots,{\bm h}_M). \label{eq:graph_representation}
\end{equation}

The graph-level representation is subsequently utilized to enhance energy-based OOD detection through cross-entropy loss $\mathcal L_{cls}$ and energy regularization loss $\mathcal L_{energy}$:
\begin{small}
\begin{equation}
\underset{\theta \in \Theta}{\min} \mathbb{E}_{(G,y)\sim\mathcal{D}_{tr}}
\big[\underbrace{-\log f_y(G)}_{\mathcal L_{cls}}
+
\lambda \cdot \underbrace{ \text{ReLU}\big(E(G)\!-\!m_{in}\big)\!^{2} }_{\mathcal L_{energy}}\big],
\label{newnewnew}
\end{equation}
\end{small}
where \(E(G;f)=-T\cdot{ \log} \sum_{i=1}^{N}e^{f_i(G)/ T}\), and \(m_{in}\) represents the  margins hyperparameter.  During inference, the test image is first split into patches by the pre-trained ViT, and subsequently processed by the well-trained ViG to compute the final confidence score for OOD detection.
\section{Experiments}

\subsection{Experimental Details}
\subsubsection{\textbf{Datasets.}}We conduct  experiments on two benchmarks:
\begin{itemize}[leftmargin=*,topsep=5pt]
\item {The small-scale \texttt{CIFAR-100} \cite{krizhevsky2009learning}. Following common practice \cite{huang2021mos}, the OOD datasets used are \texttt{CIFAR-10} \cite{krizhevsky2009learning}, \texttt{SVHN} \cite{netzer2011reading}, \texttt{Texture} \cite{cimpoi2014describing} and \texttt{Places365} \cite{zhou2017places}.}

\item {The large-scale  \texttt{ImageNet-1K} \cite{deng2009imagenet}. The OOD datasets used are \texttt{iNaturalist} \cite{van2018inaturalist}, \texttt{SUN} \cite{xiao2010sun}, \texttt{Places} \cite{zhou2017places}, and \texttt{Texture} \cite{cimpoi2014describing}.}

\end{itemize}
\subsubsection{\textbf{Baselines.}} We compare our method with two types of baselines: zero-shot methods and methods that require additional training. For zero-shot methods, we select Energy \cite{liu2020energy}, MCM \cite{ming2022delving}, CLIPN \cite{wang2023clipn}, Neglabel \cite{jiang2024negative}, among others. For methods that require training,  we compare with  MaxLogit \cite{hendrycks2019scaling}, CoOp \cite{zhou2022coop}, NegPrompt \cite{li2024learning}, etc.  Note that global pooling and subsequent training are applied exclusively to the aggregated image representations, with no textual features involved. More details are in Baselines section in the appendix.

\begin{table*}[ht!]
\captionsetup{skip=5pt}
\caption{OOD detection results on the small-scale \texttt{CIFAR-100} dataset, in terms of AUROC, AUPR, and FPR95 (mean \(\pm\) std). The best results are highlighted in \textbf{bold}, while the second-best results are \underline{underlined}. \(\uparrow\) (\(\downarrow\)) indicates that the larger (smaller) values are better. The first section are zero-shot methods, while the second section are training-required methods.}
\centering
\scriptsize
\setlength{\tabcolsep}{2pt}
\renewcommand{\arraystretch}{0.85}
\begin{tabular}{lcccccccccccc|c}
\toprule
\multirow{2}{*}{\textbf{}} & \multicolumn{3}{c}{\texttt{CIFAR-10}} & \multicolumn{3}{c}{\texttt{SVHN}} & \multicolumn{3}{c}{\texttt{Texture}} & \multicolumn{3}{c|}{\texttt{Places365}} & \multirow{2.5}{*}{ID-Acc$\uparrow$}\\
\cmidrule(lr){2-4}\cmidrule(lr){5-7}\cmidrule(lr){8-10}\cmidrule(lr){11-13}
& AUROC$\uparrow$ & AUPR$\uparrow$ & FPR95$\downarrow$ & AUROC$\uparrow$ & AUPR$\uparrow$ & FPR95$\downarrow$ & AUROC$\uparrow$ & AUPR$\uparrow$ & FPR95$\downarrow$  & AUROC$\uparrow$ & AUPR$\uparrow$ & FPR95$\downarrow$ &  \\
\midrule 
% \multicolumn{14}{c}{\textbf{Zero-shot Methods}} \\
% \midrule
%\cite{liu2020energy}
Energy  & \scriptsize{84.04}$\pm$\tiny{1.12} & \scriptsize{82.72}$\pm$\tiny{1.36} & \scriptsize{59.16}$\pm$\tiny{2.07} & \scriptsize{88.20}$\pm$\tiny{2.60} & \scriptsize{89.72}$\pm$\tiny{1.82} & \scriptsize{72.54}$\pm$\tiny{2.68} & \scriptsize{80.43}$\pm$\tiny{1.94} & \scriptsize{85.12}$\pm$\tiny{2.73} & \scriptsize{65.55}$\pm$\tiny{0.45} & \scriptsize{83.47}$\pm$\tiny{1.45} & \scriptsize{80.45}$\pm$\tiny{1.68} & \scriptsize{59.86}$\pm$\tiny{2.53} & \scriptsize{78.30}$\pm$\tiny{0.87} \\
%\cite{ming2022delving}
MCM  & \scriptsize{83.08}$\pm$\tiny{2.31} & \scriptsize{73.41}$\pm$\tiny{1.74} & \scriptsize{78.36}$\pm$\tiny{2.14} & \scriptsize{89.96}$\pm$\tiny{2.12} & \scriptsize{88.72}$\pm$\tiny{0.20} & \scriptsize{64.45}$\pm$\tiny{2.77} & \scriptsize{73.61}$\pm$\tiny{0.59} & \scriptsize{82.10}$\pm$\tiny{2.01} & \scriptsize{90.30}$\pm$\tiny{1.34} & \scriptsize{61.37}$\pm$\tiny{0.92} & \scriptsize{60.91}$\pm$\tiny{1.35} & \scriptsize{98.42}$\pm$\tiny{2.28} & \scriptsize{76.34}$\pm$\tiny{1.08} \\
%\cite{wang2023clipn}
CLIPN  & \scriptsize{88.66}$\pm$\tiny{1.08} & \scriptsize{89.12}$\pm$\tiny{0.14} & \scriptsize{50.67}$\pm$\tiny{2.31} & \scriptsize{88.20}$\pm$\tiny{0.17} & \scriptsize{49.82}$\pm$\tiny{0.14} & \scriptsize{71.72}$\pm$\tiny{2.82} & \scriptsize{90.92}$\pm$\tiny{1.65} & \scriptsize{93.38}$\pm$\tiny{1.89} & \scriptsize{37.74}$\pm$\tiny{0.91} & \scriptsize{87.25}$\pm$\tiny{1.71} & \scriptsize{86.16}$\pm$\tiny{0.94} & \scriptsize{51.06}$\pm$\tiny{1.87} & \scriptsize{79.62}$\pm$\tiny{0.75} \\
%\cite{jiang2024negative}
Neglabel  & \scriptsize{77.60}$\pm$\tiny{1.34} & \scriptsize{78.34}$\pm$\tiny{1.39} & \scriptsize{72.09}$\pm$\tiny{2.17} & \scriptsize{93.15}$\pm$\tiny{0.91} & \scriptsize{86.81}$\pm$\tiny{1.32} & \underline{\scriptsize{22.78}}$\pm$\tiny{2.49} & \scriptsize{90.40}$\pm$\tiny{1.45} & \scriptsize{91.31}$\pm$\tiny{2.47} & \scriptsize{56.11}$\pm$\tiny{1.08} & \scriptsize{89.74}$\pm$\tiny{1.79} & \scriptsize{79.35}$\pm$\tiny{0.97} & \scriptsize{40.86}$\pm$\tiny{1.52} & \scriptsize{76.20}$\pm$\tiny{0.93} \\
 
\midrule 
% \multicolumn{14}{c}{\textbf{Zero-shot Methods}} \\
% \midrule
% \multicolumn{14}{c}{\textbf{Training-required Methods}} \\
% \midrule
%\cite{hendrycks2016baseline}
MSP  & \scriptsize{78.31}$\pm$\tiny{1.21} & \scriptsize{79.52}$\pm$\tiny{1.32} & \scriptsize{81.82}$\pm$\tiny{2.03} & \scriptsize{76.04}$\pm$\tiny{1.12} & \scriptsize{60.76}$\pm$\tiny{1.10} & \scriptsize{83.69}$\pm$\tiny{0.94} & \scriptsize{76.93}$\pm$\tiny{1.23} & \scriptsize{85.24}$\pm$\tiny{0.76} & \scriptsize{83.83}$\pm$\tiny{1.42} & \scriptsize{79.44}$\pm$\tiny{1.38} & \scriptsize{62.39}$\pm$\tiny{1.41} & \scriptsize{81.24}$\pm$\tiny{1.79} & \scriptsize{77.13}$\pm$\tiny{0.83} \\
 %\cite{liang2017enhancing}
ODIN  & \scriptsize{78.18}$\pm$\tiny{1.13} & \scriptsize{79.12}$\pm$\tiny{1.41} & \scriptsize{83.16}$\pm$\tiny{1.92} & \scriptsize{71.08}$\pm$\tiny{2.85} & \scriptsize{52.36}$\pm$\tiny{1.37} & \scriptsize{89.76}$\pm$\tiny{1.56} & \scriptsize{79.39}$\pm$\tiny{2.01} & \scriptsize{86.67}$\pm$\tiny{1.35} & \scriptsize{78.37}$\pm$\tiny{2.58} & \scriptsize{79.83}$\pm$\tiny{2.01} & \scriptsize{60.85}$\pm$\tiny{1.93} & \scriptsize{81.27}$\pm$\tiny{0.74} & 7\scriptsize{6.92}$\pm$\tiny{2.91} \\
 %\cite{huang2021importance}
GradNorm  & \scriptsize{71.33}$\pm$\tiny{0.97} & \scriptsize{67.28}$\pm$\tiny{2.14} & \scriptsize{82.32}$\pm$\tiny{2.52} & \scriptsize{71.32}$\pm$\tiny{2.20} & \scriptsize{50.77}$\pm$\tiny{2.36} & \scriptsize{79.72}$\pm$\tiny{1.64} & \scriptsize{64.75}$\pm$\tiny{2.34} & \scriptsize{70.58}$\pm$\tiny{1.97} & \scriptsize{82.15}$\pm$\tiny{1.83} & \scriptsize{69.64}$\pm$\tiny{1.67} & \scriptsize{36.36}$\pm$\tiny{0.82} & \scriptsize{81.98}$\pm$\tiny{2.58} & \scriptsize{77.99}$\pm$\tiny{1.04} \\
 %\cite{sun2021react}
ReAct  & \scriptsize{76.62}$\pm$\tiny{1.35} & \scriptsize{78.97}$\pm$\tiny{0.23} & \scriptsize{70.81}$\pm$\tiny{2.25} & \scriptsize{83.73}$\pm$\tiny{1.80} & \scriptsize{76.43}$\pm$\tiny{0.60} & \scriptsize{77.41}$\pm$\tiny{0.55} & \scriptsize{81.73}$\pm$\tiny{1.45} & \scriptsize{89.01}$\pm$\tiny{2.29} & \scriptsize{76.76}$\pm$\tiny{0.67} & \scriptsize{79.63}$\pm$\tiny{2.44} & \scriptsize{59.44}$\pm$\tiny{1.79} & \scriptsize{79.18}$\pm$\tiny{0.96} & \scriptsize{75.80}$\pm$\tiny{0.94} \\
 %\cite{wang2022vim}
VIM  & \scriptsize{71.50}$\pm$\tiny{1.42} & \scriptsize{71.39}$\pm$\tiny{1.65} & \scriptsize{88.00}$\pm$\tiny{1.84} & \scriptsize{81.20}$\pm$\tiny{0.47} & \scriptsize{72.82}$\pm$\tiny{1.54} & \scriptsize{82.79}$\pm$\tiny{2.91} & \scriptsize{87.41}$\pm$\tiny{0.88} & \scriptsize{92.15}$\pm$\tiny{1.14} & \scriptsize{55.90}$\pm$\tiny{2.36} & \scriptsize{75.76}$\pm$\tiny{0.85} & \scriptsize{56.24}$\pm$\tiny{2.07} & \scriptsize{83.85}$\pm$\tiny{1.35} & \scriptsize{72.31}$\pm$\tiny{1.15} \\
 %\cite{sun2022out}
KNN  & \scriptsize{76.52}$\pm$\tiny{2.26} & \scriptsize{73.01}$\pm$\tiny{1.82} & \scriptsize{82.11}$\pm$\tiny{2.16} & \scriptsize{82.21}$\pm$\tiny{0.59} & \scriptsize{71.46}$\pm$\tiny{1.78} & \scriptsize{74.27}$\pm$\tiny{2.33} & \scriptsize{83.81}$\pm$\tiny{2.73} & \scriptsize{89.44}$\pm$\tiny{0.63} & \scriptsize{66.40}$\pm$\tiny{1.87} & \scriptsize{79.10}$\pm$\tiny{1.52} & \scriptsize{57.47}$\pm$\tiny{1.74} & \scriptsize{78.74}$\pm$\tiny{2.36} & \scriptsize{69.92}$\pm$\tiny{1.24} \\
 %\cite{hendrycks2019scaling}
MaxLogit  & \scriptsize{85.07}$\pm$\tiny{1.29} & \scriptsize{78.13}$\pm$\tiny{1.46} & \scriptsize{57.58}$\pm$\tiny{1.93} & \scriptsize{91.01}$\pm$\tiny{1.80} & \scriptsize{90.14}$\pm$\tiny{0.51} & \scriptsize{59.05}$\pm$\tiny{1.79} & \scriptsize{82.08}$\pm$\tiny{1.31} & \scriptsize{87.93}$\pm$\tiny{1.72} & \scriptsize{62.82}$\pm$\tiny{2.06} & \scriptsize{80.88}$\pm$\tiny{2.08} & \scriptsize{69.48}$\pm$\tiny{0.79} & \scriptsize{65.58}$\pm$\tiny{1.62} & \scriptsize{79.48}$\pm$\tiny{0.96} \\
 
 %\cite{cao2024envisioning}
EOE  & \scriptsize{85.67}$\pm$\tiny{1.24} & \scriptsize{80.32}$\pm$\tiny{0.37} & \scriptsize{59.17}$\pm$\tiny{2.06} & \scriptsize{88.78}$\pm$\tiny{0.06} & \scriptsize{86.10}$\pm$\tiny{2.85} & \scriptsize{68.47}$\pm$\tiny{0.27} & \scriptsize{82.64}$\pm$\tiny{2.58} & \scriptsize{90.31}$\pm$\tiny{1.39} & \scriptsize{66.89}$\pm$\tiny{0.79} & \scriptsize{78.06}$\pm$\tiny{2.13} & \scriptsize{61.14}$\pm$\tiny{1.59} & \scriptsize{77.60}$\pm$\tiny{0.81} & \scriptsize{78.82}$\pm$\tiny{0.85} \\
 %\cite{du2022unknown}
VOS  & \scriptsize{79.23}$\pm$\tiny{1.16} & \scriptsize{80.03}$\pm$\tiny{1.28} & \scriptsize{58.87}$\pm$\tiny{2.12} & \scriptsize{85.01}$\pm$\tiny{2.91} & \scriptsize{74.34}$\pm$\tiny{2.90} & \scriptsize{47.44}$\pm$\tiny{0.59} & \scriptsize{78.26}$\pm$\tiny{0.84} & \scriptsize{85.56}$\pm$\tiny{1.57} & \scriptsize{61.36}$\pm$\tiny{2.41} & \scriptsize{79.71}$\pm$\tiny{1.26} & \scriptsize{61.41}$\pm$\tiny{2.41} & \scriptsize{57.17}$\pm$\tiny{1.64} & \scriptsize{77.04}$\pm$\tiny{2.91} \\
 %\cite{zhou2022learning}
CoOp  & \scriptsize{76.01}$\pm$\tiny{1.47} & \scriptsize{74.12}$\pm$\tiny{1.52} & \scriptsize{67.30}$\pm$\tiny{2.43} & \scriptsize{87.97}$\pm$\tiny{2.50} & \scriptsize{84.79}$\pm$\tiny{2.43} & \scriptsize{39.89}$\pm$\tiny{0.14} & \scriptsize{69.95}$\pm$\tiny{1.67} & \scriptsize{86.34}$\pm$\tiny{0.29} & \scriptsize{86.36}$\pm$\tiny{1.75} & \scriptsize{56.78}$\pm$\tiny{0.69} & \scriptsize{56.34}$\pm$\tiny{1.97} & \scriptsize{91.63}$\pm$\tiny{2.51} & \scriptsize{72.83}$\pm$\tiny{1.12} \\
 %\cite{zhou2022conditional}
CoCoOp & \scriptsize{76.85}$\pm$\tiny{1.33} & \scriptsize{74.98}$\pm$\tiny{2.56} & \scriptsize{65.73}$\pm$\tiny{2.35} & \scriptsize{90.50}$\pm$\tiny{0.64} & \scriptsize{88.96}$\pm$\tiny{0.91} & \scriptsize{31.80}$\pm$\tiny{0.98} & \scriptsize{71.55}$\pm$\tiny{1.28} & \scriptsize{84.19}$\pm$\tiny{1.49} & \scriptsize{85.80}$\pm$\tiny{2.68} & \scriptsize{58.45}$\pm$\tiny{1.94} & \scriptsize{54.19}$\pm$\tiny{2.58} & \scriptsize{91.37}$\pm$\tiny{1.72} & \scriptsize{71.97}$\pm$\tiny{1.05} \\
 %\cite{miyai2023locoop}
LoCoOp  & \scriptsize{77.67}$\pm$\tiny{2.28} & \scriptsize{76.24}$\pm$\tiny{1.49} & \scriptsize{83.61}$\pm$\tiny{1.19} & \scriptsize{93.89}$\pm$\tiny{0.55} & \scriptsize{92.39}$\pm$\tiny{0.29} & \scriptsize{29.27}$\pm$\tiny{1.17} & \scriptsize{78.61}$\pm$\tiny{0.42} & \scriptsize{93.52}$\pm$\tiny{2.06} & \scriptsize{71.17}$\pm$\tiny{1.57} & \scriptsize{48.08}$\pm$\tiny{1.35} & \scriptsize{48.20}$\pm$\tiny{0.89} & \scriptsize{97.59}$\pm$\tiny{1.47} & \scriptsize{72.34}$\pm$\tiny{1.11} \\
 %\cite{li2024learning}
NegPrompt  & \scriptsize{88.53}$\pm$\tiny{1.09} & \scriptsize{89.01}$\pm$\tiny{1.18} & \scriptsize{49.96}$\pm$\tiny{2.22} & \scriptsize{94.19}$\pm$\tiny{0.55} & \scriptsize{92.18}$\pm$\tiny{2.05} & \scriptsize{30.99}$\pm$\tiny{0.81} & \scriptsize{91.14}$\pm$\tiny{0.92} & \scriptsize{93.11}$\pm$\tiny{1.79} & \scriptsize{24.13}$\pm$\tiny{0.84} & \scriptsize{88.34}$\pm$\tiny{2.02} & \scriptsize{90.16}$\pm$\tiny{1.54} & \scriptsize{44.31}$\pm$\tiny{2.65} & \scriptsize{78.64}$\pm$\tiny{0.89} \\
 
\midrule
 
{\sysname}-DS & \scriptsize{89.06}$\pm$\tiny{1.07} & \scriptsize{90.16}$\pm$\tiny{1.11} &
\scriptsize{60.86}$\pm$\tiny{2.31} &
\scriptsize{94.88}$\pm$\tiny{0.87} & 
\underline{\scriptsize{94.72}}$\pm$\tiny{0.10} &
\scriptsize{30.37}$\pm$\tiny{1.63} & 
\scriptsize{95.18}$\pm$\tiny{0.95} & 
\textbf{\scriptsize{98.98}}$\pm$\tiny{2.48} &
\scriptsize{22.39}$\pm$\tiny{1.64} &
\scriptsize{90.23}$\pm$\tiny{0.94} &
\scriptsize{89.99}$\pm$\tiny{1.62} & 
\textbf{\scriptsize{28.62}}$\pm$\tiny{2.55} &
\scriptsize{80.96}$\pm$\tiny{1.88} \\
 
{\sysname}-Gem & \scriptsize{86.65}$\pm$\tiny{1.14} & 87.70$\pm$\tiny{1.22} &
\scriptsize{60.15}$\pm$\tiny{2.14} &
\scriptsize{94.86}$\pm$\tiny{1.84} &
\scriptsize{94.36}$\pm$\tiny{2.73} &
\scriptsize{29.77}$\pm$\tiny{0.42} & 
\underline{\scriptsize{95.32}}$\pm$\tiny{1.28} &
\scriptsize{98.91}$\pm$\tiny{1.56} & 
\underline{\scriptsize{20.58}}$\pm$\tiny{0.71} &
\underline{\scriptsize{92.84}}$\pm$\tiny{2.08} & \underline{\scriptsize{92.69}}$\pm$\tiny{1.35} & \scriptsize{33.42}$\pm$\tiny{1.46} & \scriptsize{82.03}$\pm$\tiny{0.81} \\
 
{\sysname}-OAI & 
\textbf{\scriptsize{89.72}}$\pm$\tiny{0.97} & \underline{\scriptsize{90.33}}$\pm$\tiny{1.06} & \textbf{\scriptsize{49.08}}$\pm$\tiny{1.88} & \underline{\scriptsize{95.12}}$\pm$\tiny{0.42} &
\scriptsize{94.65}$\pm$\tiny{0.78} & \scriptsize{27.84}$\pm$\tiny{2.41} & \scriptsize{93.64}$\pm$\tiny{2.48} & \scriptsize{98.48}$\pm$\tiny{0.66} & \scriptsize{28.19}$\pm$\tiny{1.89} & \scriptsize{92.42}$\pm$\tiny{1.78} & \scriptsize{92.62}$\pm$\tiny{2.09} & \scriptsize{35.70}$\pm$\tiny{0.93} & \scriptsize{80.93}$\pm$\tiny{0.86} \\
 
{\sysname}-Clau & \scriptsize{87.69}$\pm$\tiny{1.10} & \scriptsize{86.70}$\pm$\tiny{0.96} & \scriptsize{55.34}$\pm$\tiny{2.25} & \scriptsize{94.30}$\pm$\tiny{0.88} & \scriptsize{93.49}$\pm$\tiny{1.99} & \scriptsize{30.99}$\pm$\tiny{0.22} & \scriptsize{94.44}$\pm$\tiny{1.38} & \scriptsize{98.71}$\pm$\tiny{2.14} & \scriptsize{29.78}$\pm$\tiny{0.92} & \scriptsize{91.80}$\pm$\tiny{1.26} & \scriptsize{91.84}$\pm$\tiny{2.54} & \scriptsize{35.81}$\pm$\tiny{1.39} & \scriptsize{82.03}$\pm$\tiny{0.81} \\
{\sysname}-GPT & \underline{\scriptsize{89.63}}$\pm$\tiny{0.98} & \textbf{\scriptsize{90.34}}$\pm$\tiny{1.05} & \underline{\scriptsize{49.75}}$\pm$\tiny{2.31} & \textbf{\scriptsize{96.29}}$\pm$\tiny{1.30} & \textbf{\scriptsize{95.97}}$\pm$\tiny{1.49} & \textbf{\scriptsize{19.92}}$\pm$\tiny{0.84} & \textbf{\scriptsize{95.55}}$\pm$\tiny{1.74} & \underline{\scriptsize{98.94}}$\pm$\tiny{1.09} & \textbf{\scriptsize{18.83}}$\pm$\tiny{1.56} & \textbf{\scriptsize{93.71}}$\pm$\tiny{1.47} & \textbf{\scriptsize{93.56}}$\pm$\tiny{2.41} & \underline{\scriptsize{30.23}}$\pm$\tiny{1.73} & \scriptsize{78.53}$\pm$\tiny{1.34} \\
\bottomrule
\addlinespace[3pt]
\multicolumn{12}{c}{{\tiny{Note: Due to space constraints, we abbreviate LLMs as follows: DeepSeek (DS), Gemini (Gem), OpenAI o1 (OAI), Claude  (Clau), and GPT-4o (GPT).}}}
\vspace{-0.35cm}
\label{tabel_1}
\end{tabular}
\end{table*}

%, \lambda^-_2=0.00001

\begin{figure}[t!]
\centering
\captionsetup{skip=5pt}
\includegraphics[width=0.68\linewidth]{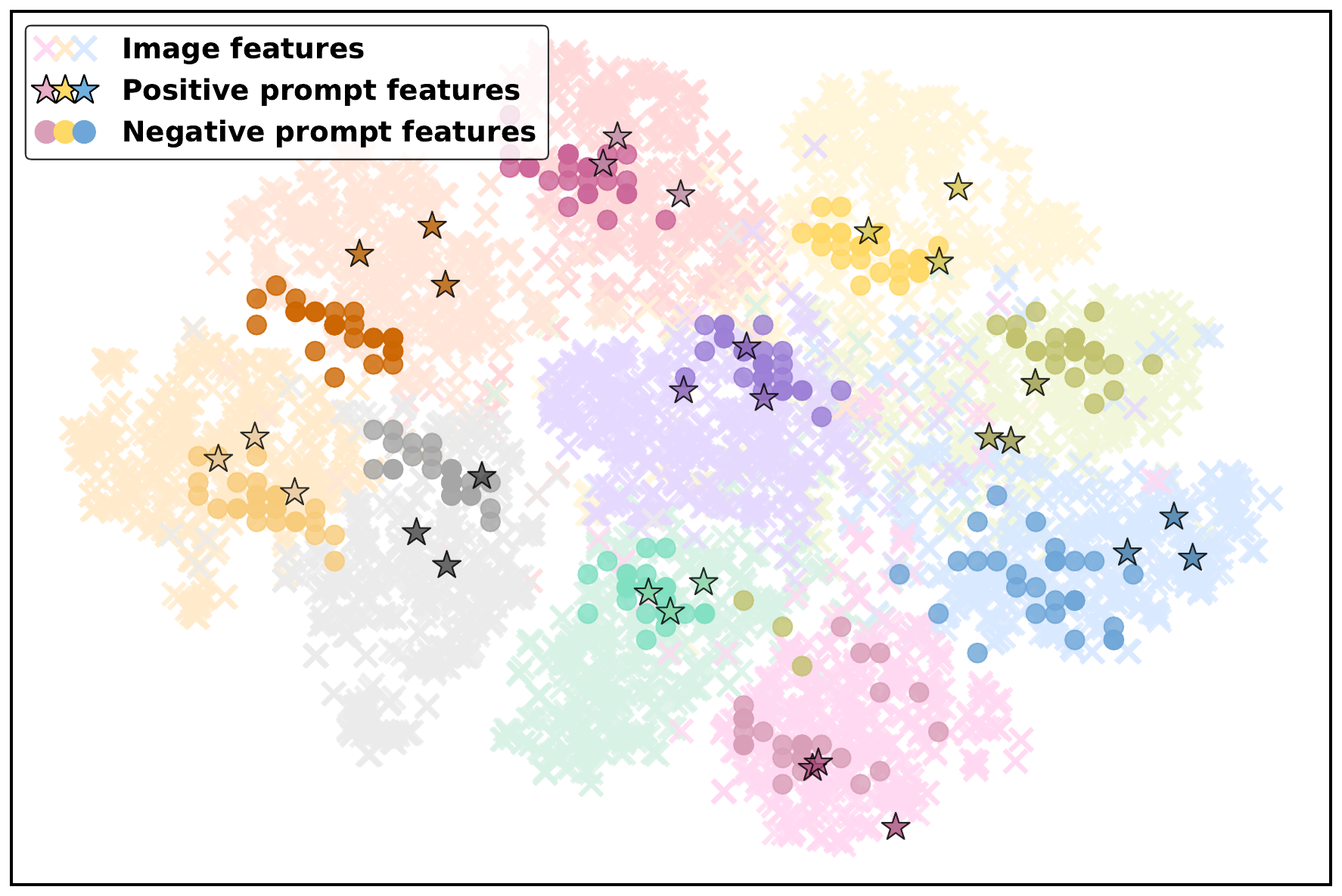}
\caption{T-SNE visualization of optimized image features, along with positive and negative prompt features on the \texttt{CIFAR-10}  dataset in the shared semantic space.}
\label{tsne}
\vspace{-0.5cm}
\end{figure}

\subsubsection{\textbf{Experimental Settings.}} All experiments are conducted using the pre-trained CLIP (ViT-B/16) as the backbone, with encoder parameters frozen during training. For category feature generation, we employ several commonly used LLMs: GPT-4o, DeepSeek-R1, Gemini 2.0 Pro, OpenAI o1, and Claude 3.7 Sonnet. The hyperparameters for the positive and negative losses, $\lambda^+$ and $\lambda^-$, are set to $1e-5$ and $1e-3$, respectively. The number of super-classes is set to $20$ for \texttt{CIFAR-100} and $50$ for \texttt{ImageNet-1K}. For the ViG model, we use the isotropic architecture with $4$ and $5$ interactive GCN layers for the two datasets, respectively. The Top-K values $\{{K}^T,{K}^P,{K}^M\}$ are set to $\{{2, 10, 8}\}$ for \texttt{CIFAR-100} and $\{{2, 20, 18}\}$ for \texttt{ImageNet-1K}. The margin $m_{in}$ is set to $10$ and $12$ for two datasets. All experiments are performed on two NVIDIA A800 GPUs. Following common practice \cite{huang2021mos}, we evaluate our method using AUROC, AUPR, FPR95, and ID-Acc.

\subsection{Experimental Results}
We report the OOD detection results on the \texttt{CIFAR-100} dataset in Table \ref{tabel_1}. The average AUROC scores for five different LLMs are $92.34\%$, $92.42\%$, $92.73\%$, $92.06\%$, and $93.80\%$, respectively. These results are consistent, with a minor variance of  $3e-5$, suggesting that the choice of LLMs has little impact on overall performance, and that subsequent optimization and training steps are more critical. Notably, GPT-4o achieves the best overall performance among the five LLMs, underscoring its relative advantage in text understanding and the generation of discriminative content. Among zero-shot baselines, while CLIPN achieves competitive results by leveraging negative prompts to empower the logic of saying ``no" within CLIP, it may introduce overlapping and noisy OOD features. In contrast, our method guides negative prompts to focus on ID features near category boundaries, resulting in higher AUROC, with improvements of  $0.97\%$ on \texttt{CIFAR-10}, $8.09\%$ on \texttt{SVHN}, $4.63\%$ on \texttt{Texture}, and $6.46\%$ on \texttt{Places365} over CLIPN.

\begin{figure}[t!]
\setlength{\abovecaptionskip}{2pt}
    \centering
    \captionsetup{skip=5pt}
\includegraphics[width=0.48\textwidth,height=2.7cm]{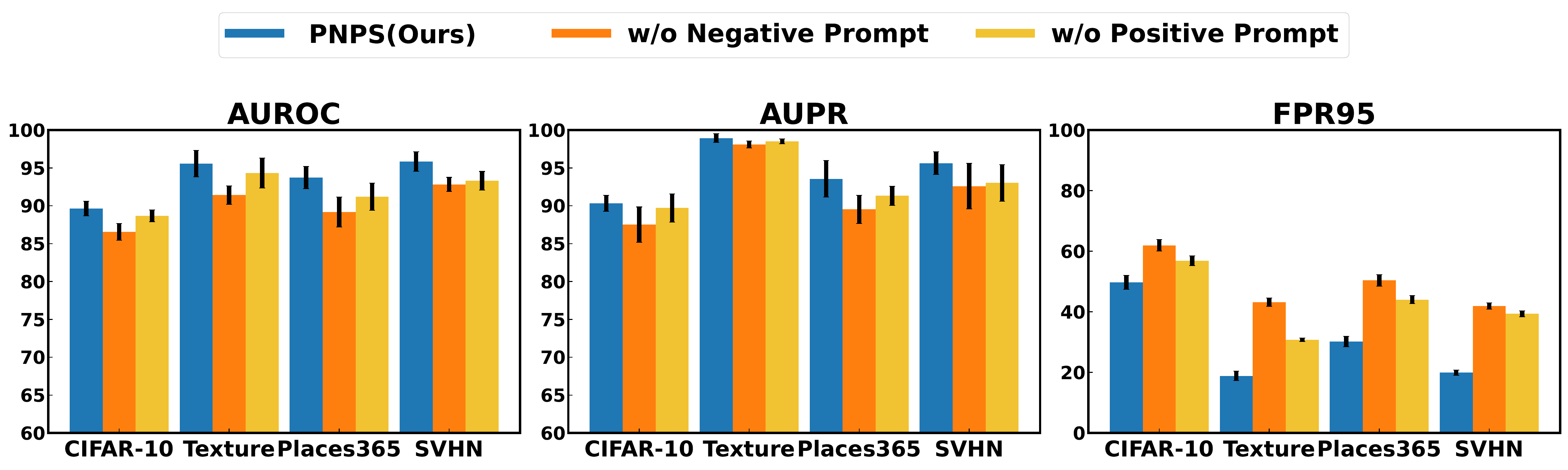}
    \caption{Performance in terms of AUROC, AUPR, and FPR95 under different settings on \texttt{CIFAR-100} dataset.}
    \label{fig:bar_chart_cifar100}
    \vspace{-0.6cm}
\end{figure}

\begin{figure*}[t!]
\centering
\captionsetup{skip=5pt}
\includegraphics[width=0.92\linewidth]{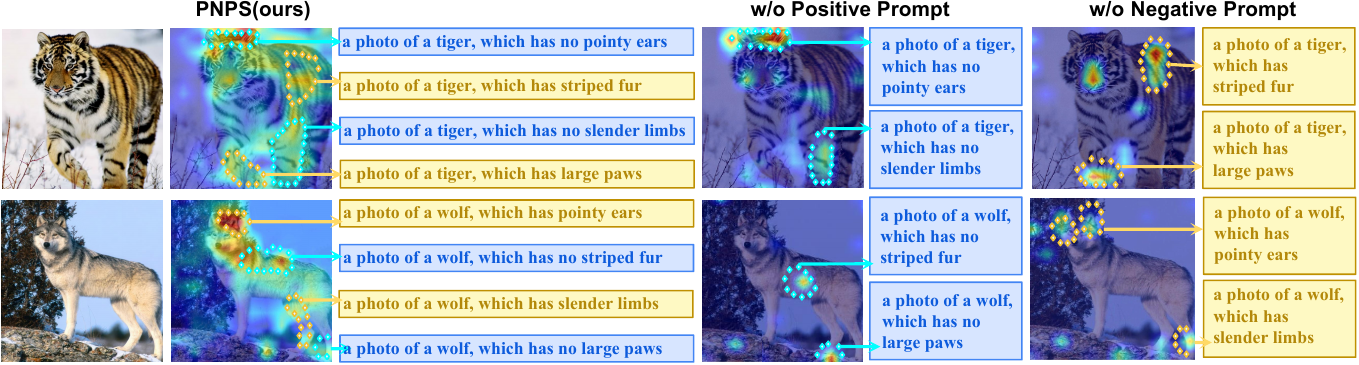}
\caption{Grad-CAM maps visualization of features captured by full, without positive,  and without negative prompts. }
\vspace{-0.4cm}
\label{gradcam}
\end{figure*}

Among the training-based baselines, NegPrompt, which captures negative semantics relative to ID classes, achieves the best overall performance. On the \texttt{CIFAR-10}, \texttt{SVHN}, \texttt{Texture}, and \texttt{Places365} datasets, our {\sysname} outperforms it in terms of AUROC by $1.19\%$, $1.65\%$, $4.41\%$, and $5.37\%$, respectively. While our model achieves comparable performance to NegPrompt on the more challenging datasets (\texttt{CIFAR-10} and \texttt{SVHN}), it demonstrates a substantially greater advantage on the simpler datasets (\texttt{Texture} and \texttt{Places365}). These results indicate that our method not only maintains robustness in difficult scenarios but also excels in less complex tasks. The results  for \texttt{ImageNet-1K} are provided in Experimental Results in the appendix.

\begin{figure}[t!]
\setlength{\abovecaptionskip}{2pt}
    \centering
    \begin{subfigure}[t]{0.49\linewidth}
        \centering
        \captionsetup{skip=5pt}
        \includegraphics[width=\linewidth,height=3cm]{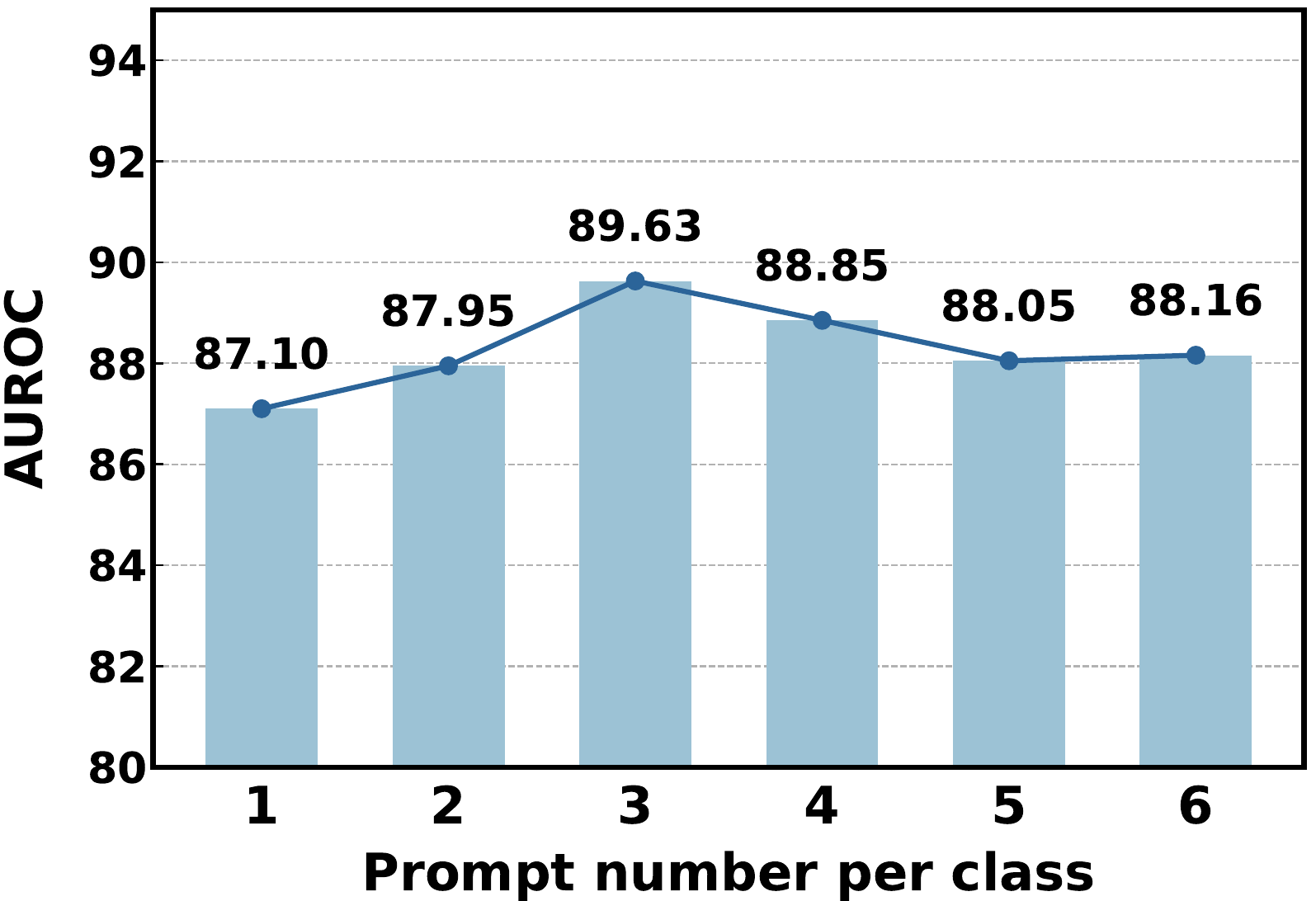}
        \subcaption{AUROC for feature count.}
        \label{picturea}
    \end{subfigure}
    \hfill
    \begin{subfigure}[t]{0.49\linewidth}
        \centering
        \captionsetup{skip=5pt}
        \includegraphics[width=\linewidth,height=3cm]{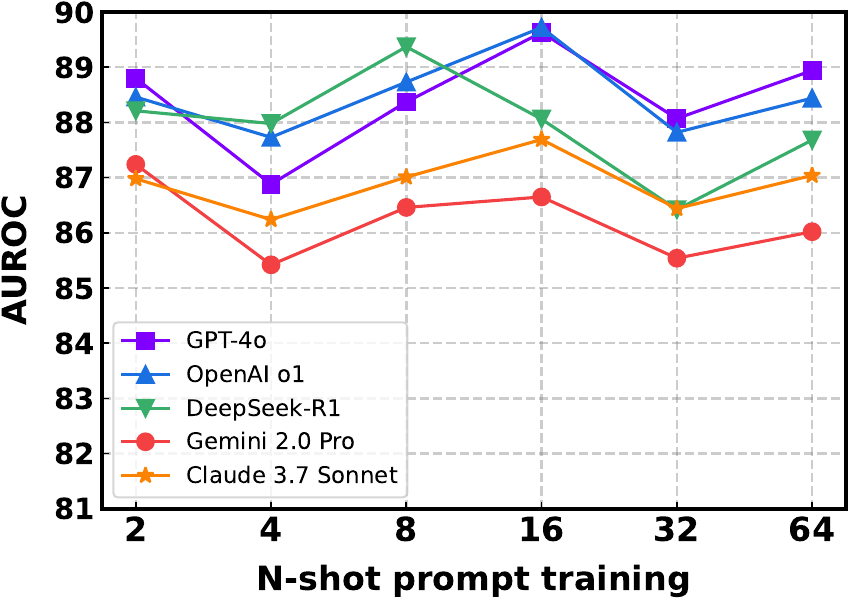}
        \subcaption{AUROC for N-shot learning.}
        \label{pictureb}
    \end{subfigure}
    \captionsetup{skip=5pt}
    \caption{Left: AUROC for LLMs-generated feature count. Right: AUROC for prompt learning under N-shot settings.}
    \label{fig:main}
    \vspace{-0.5cm}
\end{figure}

\subsection{\textbf{Experimental Analysis}}
For simplicity, we use the same configuration to construct and optimize prompts on the \texttt{CIFAR-10} dataset. The image representations, together with the positive and negative prompt representations, are shown in Figure \ref{tsne}.  The visualization reveals that image representations from different categories form well-separated clusters. Notably, positive representations are mainly concentrated within each class, while negative ones are distributed along class boundaries, further enhancing inter-class separability. These negative representations not only complement the limited positive representations but also reinforce inter-class boundaries, enabling the classifier to learn clearer decision boundaries.

Following the above visualization in the shared semantic space, we further analyze the specific visual features learned by the optimized positive and negative prompts for a given image. As shown in Figure \ref{gradcam}, we present Grad-CAM visualization of the image regions highlighted by both prompts, as well as by negative prompts alone and positive prompts alone. It is evident that combining both prompts enables the model to focus on a broader range of feature regions. Moreover, the visual regions highlighted by the prompts align well with their textual semantics, demonstrating the effectiveness of our optimization. Specifically, positive prompts tend to focus on distinctive category features, such as the striped fur of a tiger, whereas negative prompts emphasize the absence of these distinctive features in other categories, thereby highlighting inter-class differences.

\subsection{Ablation Study}
\subsubsection{The Effectiveness of Negative and Positive Prompts.} We conduct ablation experiments to evaluate the contribution of positive and negative prompt representations by removing each component individually.  The conditions ``w/o positive prompt" and ``w/o negative prompt" illustrate the impact of excluding positive and negative  representations, as shown in Figure \ref{fig:bar_chart_cifar100}. Our results demonstrate that retaining only negative representations yields better performance than retaining only positive representations.
This finding, together with the t-SNE visualization in Figure \ref{tsne}, further validates that negative representations are more effective than positive representations in assisting the classifier to define decision boundaries between categories for OOD detection.
Moreover, removing either module leads to performance degradation, highlighting the indispensable role of both representations in enhancing model performance.

\subsubsection{The Effectiveness Graph-Based Connection.}

Graph-based architectures inherently support information propagation between heterogeneous nodes, making them a more direct and efficient solution for modeling complex intra-modal and inter-modal relationships. To validate the effectiveness of graph-based aggregation and propagation, we directly employ the optimized positive and negative representations, together with the MCM score \cite{ming2022delving}, for VLM-based OOD detection. We conduct experiments on the \texttt{CIFAR-100} and \texttt{ImageNet-1K} datasets, and report the relevant results in the appendix. Compared to the zero-shot CLIPN method, our approach achieves slightly lower performance when the graph-based structure is omitted.  However, after incorporating graph connections, its performance improves significantly. These findings demonstrate that aggregating and propagating semantic supervision via graph-based structures can effectively enhance OOD detection.
 
% \noindent\textbf{The Effectiveness of Prompt Training.}
% As shown in Figure \ref{picturea}, we illustrate the effect of few-shot samples on the training of prompt representations for OOD detection across different large language models on the \texttt{CIFAR-100} dataset. We set the number of samples in the $N$-shot setting to $N=\{2, 4, 8, 16, 32\}$. When $N=2$, the model tends to learn basic visual semantic features, achieving competitive performance. As $N$ increases, OOD detection improves steadily. However, when $N > 16$, the model starts to overfit, capturing noisy or irrelevant features, which leads to performance degradation. The training cost for $\{2,4,8,16,32\}$ are 1.39, 3.38, 7.83, 16.32, and 30.78 $mins$ with the number of positive prompts set to 3.

\section{Conclusion}
In this work, we propose a three-phase {\sysname} framework. Initially, LLMs are leveraged to construct class-specific positive and negative prompts. Subsequently, these prompts are optimized via learnable textual matrics to capture intra-class and inter-class features, which facilitate comprehensive learning of ID features and clearer category boundaries. Additionally, a graph-based model aggregates semantic supervision from the optimized prompt representations and transfers it to image features, thus enhancing OOD detection performance. Experiments on two ID datasets and eight OOD datasets using five different LLMs demonstrate that our approach outperforms state-of-the-art baselines.

\section{Acknowledgments}
This work done by Zhixia He, Qin Tian, and Minglai Shao is supported by the National Natural Science Foundation of China (No. 62272338) and the Research Fund of the Key Lab of Education Blockchain and Intelligent Technology, Ministry of Education (EBME25-F-06). 
Chen Zhao, Xintao Wu, Xujiang Zhao, Dong Li, and Linlin Yu did not receive any financial support for this work and contributed only by developing the research ideas, participating in discussions, and providing feedback on the manuscript.

\bibliography{reference}
\newpage

\end{document}